\documentclass[a4paper]{article}
\usepackage[utf8]{inputenc}

\usepackage{INTERSPEECH2021}

\usepackage{color}
\usepackage{cite}
\usepackage{graphicx}
\usepackage{booktabs}
\usepackage{multirow}
\usepackage{tabularx}
\usepackage[hyphens]{url}
\usepackage[hidelinks]{hyperref}
\usepackage[capitalise]{cleveref}
\usepackage{enumitem}
\usepackage{stackengine}

\usepackage{units}  
\usepackage{siunitx,etoolbox}
\usepackage{tikz}
\usetikzlibrary{shapes,positioning,fit,decorations.markings,arrows, arrows.meta, automata, backgrounds}

\tikzset{
    >=stealth',
    punkt/.style={
           rectangle,
           rounded corners,
           draw=black, very thick,
           text width=6.5em,
           minimum height=2em,
           text centered},
    pil/.style={
           ->,
           thick,
           shorten <=2pt,
           shorten >=2pt,}
}
\tikzstyle{word} = [draw, top color=blue!10, bottom color=blue!50, rounded corners=1mm]
\tikzstyle{wordRed} = [draw, top color=red!10, bottom color=red!90, rounded corners=1mm]
\tikzstyle{wordGreen} = [draw, top color=green!10, bottom color=green!90, rounded corners=1mm]
\tikzstyle{wordOrange} = [draw, top color=orange!10, bottom color=orange!90, rounded corners=1mm]

\DeclareUnicodeCharacter{00A0}{}

\newcommand{\blank}{\ensuremath{\epsilon}}

\newcommand{\smalldots}{\makebox[0.8em][c]{...}}
\renewcommand{\paragraph}[1]{\textbf{#1}\hskip\parindent}

\newcommand{\decoder}{\ensuremath{\operatorname{SlowRNN}}}

\newcommand{\eos}{\ensuremath{\langle\operatorname{eos}\rangle}}

\newcommand{\encoderlength}{\ensuremath{T}}

\newcommand{\tokenlength}{\ensuremath{S}}

\newcommand{\alignpos}{\ensuremath{u}}

\newcommand{\aligncur}{\ensuremath{\alpha_u}}

\newcommand{\emitscale}{\ensuremath{\delta}}
\newcommand{\emitscaleeos}{\ensuremath{{\emitscale}_{\eos}}}
\newcommand{\lmscale}{\ensuremath{\beta}}
\newcommand{\lmscaleeos}{\ensuremath{{\lmscale}_{\eos}}}
\newcommand{\lmscaleother}{\ensuremath{{\lmscale}_{\textsc{other}}}}
\newcommand{\labelscale}{\ensuremath{\lambda}}
\newcommand{\ilmscale}{\ensuremath{\gamma}}
\newcommand{\ilmscaleother}{\ensuremath{{\ilmscale}_{\textsc{other}}}}

\newcommand\blankproblibrilhs[1]{\ensuremath{
p_{\alignpos}^{#1}(\Delta t {=} 1 \mid \smalldots)
}}

\newcommand\emitproblibrilhs[1]{\ensuremath{
p_\alignpos^{#1}(\Delta t {=} 0 \mid \smalldots)
}}

\newcommand{\ilmAvgName}{\ensuremath{\operatorname{avg}}}
\newcommand{\ilmZeroName}{\ensuremath{0}}

\DeclareMathOperator*{\argmax}{arg\,max}

\setlist{
        itemsep=0pt,
        parsep=1pt plus 1pt minus 1pt,
        topsep=1pt plus 1pt minus 1pt,
        partopsep=0pt}

\title{Librispeech Transducer Model with Internal Language Model Prior Correction}
\name{Albert Zeyer$^{1,2}$, André Merboldt$^{1}$, Wilfried Michel$^{1,2}$, Ralf Schlüter$^{1,2}$, Hermann Ney$^{1,2}$}
\address{
  $^1$Human Language Technology and Pattern Recognition,
  Computer Science Department, \\
  RWTH Aachen University, 52062 Aachen, Germany, \\
  $^2$AppTek GmbH, 52062 Aachen, Germany}
\email{\{zeyer, michel, schlueter, ney\}@cs.rwth-aachen.de, andre.merboldt@rwth-aachen.de}

\begin{document}



\maketitle
\begin{abstract}
We present our transducer model on Librispeech.
We study variants to include an external language model (LM)
with shallow fusion
and subtract an estimated internal LM.
This is justified by a Bayesian interpretation
where the transducer model prior is given by the estimated internal LM.
The subtraction of the internal LM gives us over 14\% relative improvement
over normal shallow fusion.
Our transducer has a separate probability distribution
for the non-blank labels
which allows for easier combination with the external LM,
and easier estimation of the internal LM.
We additionally take care of including the end-of-sentence (EOS) probability
of the external LM in the last blank probability
which further improves the performance.
All our code and setups are published.
\end{abstract}

\section{Introduction \& Related Work}

The recurrent neural network transducer (RNN-T) model
\cite{graves2012seqtransduction,graves2013speechrnnt}
is an end-to-end model which allows for time-synchronous decoding,
which a more natural fit for many applications such as online recognition.
Thus
RNN-T and many variations has recently gained interest
\cite{zhang2020trafotransducer,han2020contextnet,gulati2020conformer,%
variani2020hat,%
zeyer2020:transducer,%
zhou2021phonemetransducer}.

In a Bayesian interpretation,
a discriminative acoustic model $p_{\mathrm{AM}}(y \mid x)$
can be combined with an external language model $p_{\mathrm{LM}}(y)$
by
\[ p(y \mid x) =
\frac{p_{\mathrm{AM}}(y \mid x)}{ p_{\mathrm{AM}}(y) }
\cdot p_{\mathrm{AM}}(x) \cdot p_{\mathrm{LM}}(y)
\cdot \frac{1}{p(x)}. \]
In recognition, when searching for $\argmax_{y} p(y\mid x)$,
we can omit $p(x)$ and $p_{\mathrm{AM}}(x)$.
In shallow fusion, $p_{\mathrm{AM}}(y)$ is omitted as well.
In the density ratio approach \cite{mcdermott2019density},
$p_{\mathrm{AM}}(y)$ is estimated by a separate language model
trained on just the acoustic training transcriptions.
In the hybrid autoregressive transducer (HAT) \cite{variani2020hat},
$p_{\mathrm{AM}}(y)$ is estimated directly based on the implicit internal LM (ILM)
of $p_{\mathrm{AM}}(y \mid x)$.
The HAT model has a particular simple architecture
which was designed such that there is a simple approximation
for this ILM estimation by setting the encoder input to 0.
We follow up on the ILM estimation approach
and try some variations of the estimation.
Using 0 as encoder input also works
but we found some other variations to be better.

\section{Model}

\begin{figure}
\makebox[\textwidth][l]{%
\hspace{-7mm}
\includegraphics[width=1.1\columnwidth]{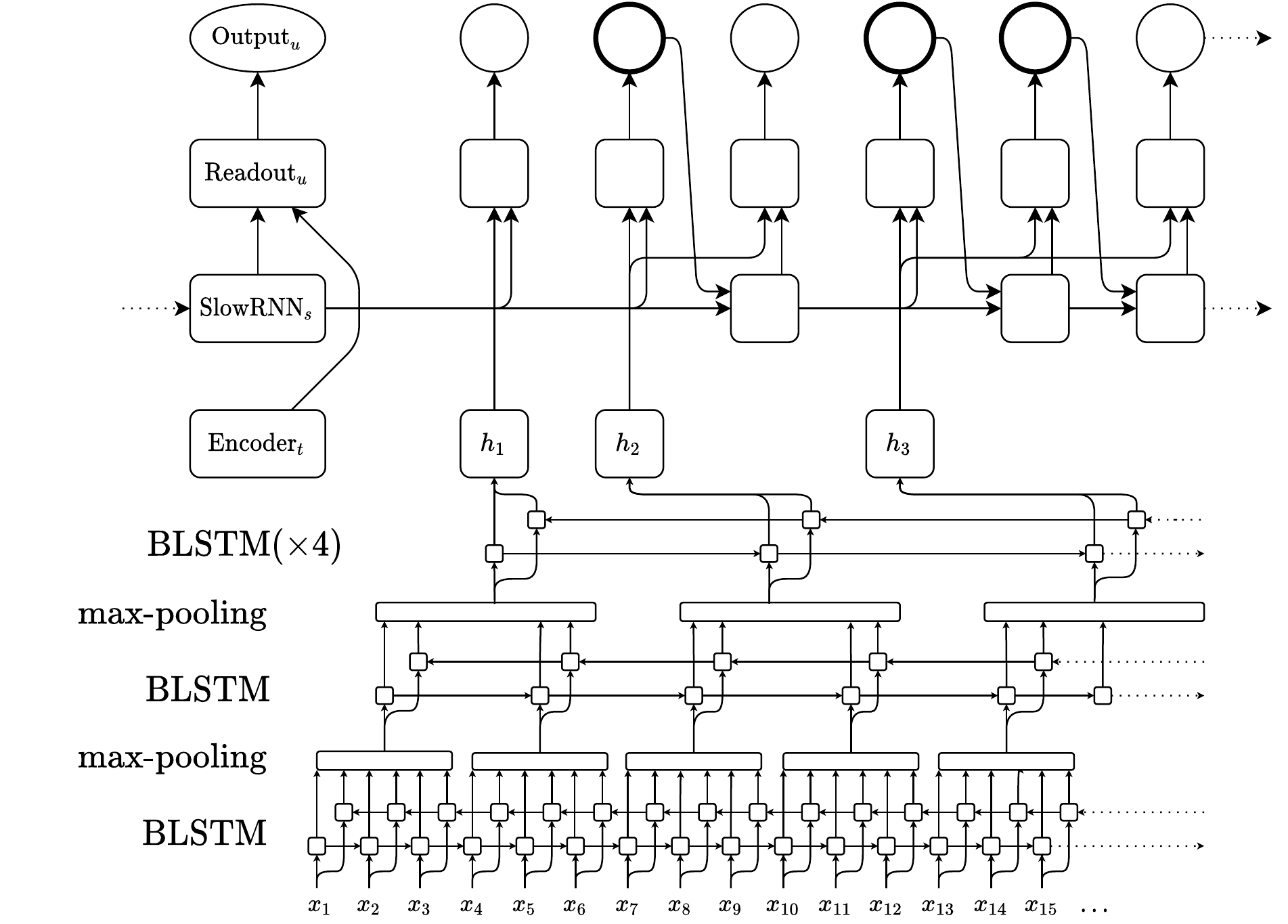}%
}
\caption[Transducer model]{Our transducer model with all dependencies.
The decoder is unrolled over the alignment axis $u$.
Compare to \cite{zeyer2020:transducer}.}
\label{fig:librispeech_transducer}
\end{figure}

We follow a transducer variant as defined in \cite{zeyer2020:transducer}.
The whole model can be seen in \Cref{fig:librispeech_transducer}.
Let $x_1^{T'}$ be the acoustic input features (MFCC in our case) of length $T'$,
and $y_1^S$ some label sequence of length $S$ over labels $\Sigma$
(excluding blank \blank).
We use \emph{byte pair encoding (BPE)}-based \emph{subword units}
\cite{sennrich2015neuralbpe,zeyer2018:asr-attention}
with a vocabulary size of about 1000 labels%
\footnote{In earlier work on attention-based encoder decoder models,
we used 10k BPE labels for Librispeech.
However, because of computation time and memory constraints,
we reduced it to 1k for the transducer model.}.

We have a multi-layer bidirectional LSTM \cite{hochreiter1997lstm} \emph{encoder} model
with interchanged max-pooling in time to downscale the input to length $T$
with factor 6.
This results in
\[ h_1^T := \operatorname{Encoder}(x_1^{T'}) . \]

We define the probability for the label sequence $y_1^S$ as
\begin{align*}
p(y_1^S \mid x_1^{T'} ) & :=
\sum_{\alpha_1^U : y_1^S} p(\alpha_1^U \mid x_1^T), \\
p(\alpha_1^U \mid x_1^T)
& := \prod_{u=1}^U p_u(\alpha_u \mid \alpha_1^{u-1}, x_1^T),
\end{align*}
with alignment label $\alpha_u \in \Sigma' := \Sigma \cup \{\blank\}$,
and where $\alpha_1^U : y_1^S$ is defined by the label topology $\mathcal{A}$.
Specifically, we use alignment length $U = T + S$
and allow all alignment label sequences $\alpha_1^U$
which match the sequence $y_1^S$ after removing all blanks \blank,
also notated as $\mathcal{A}(\alpha_1^U) = y_1^S$.
This defines an alignment between $h$ and $y$
as can be seen in \Cref{fig:lattice_rnnt}.

\begin{figure}
	\centering
	\resizebox{\columnwidth}{!}{%
\tikzstyle{fillednode}=[rectangle,draw,fill=black!20,minimum size=1.4em]
\tikzstyle{filledlabel}=[rectangle,draw,fill=red!20,minimum size=1.4em]
\tikzstyle{blankedge}=[thin,-Triangle, shorten >= 5pt, shorten <= 5pt]
\tikzstyle{labelnode} = [draw,circle,inner sep=3pt,fill]
\tikzstyle{finalnode} = [circle, inner sep=4pt, draw]

\begin{tikzpicture}

\def\maxheight{6}
\def\maxwidth{10}
\draw[blankedge] (-0.3,0) to (0,0)         node{};   
\foreach \x [evaluate={\xx=int(\x+1);}] in {0,...,\maxwidth}
{
   \draw[ultra thick] (2*\x,-0.65) -- (2*\x,-0.35) node[anchor=north] {};
  \foreach \y [evaluate={\yy=int(\y+1);}] in {0,...,\maxheight}
  {
    \draw[ultra thick] (-0.65,2*\y) -- (-0.35,2*\y) node[thick, anchor=north] {};
    \node[labelnode] at (2*\x,2*\y) {};

    \pgfmathparse{\y != \maxheight ? int(1):int(0)}
    \ifnum\pgfmathresult>0
    \draw[blankedge] (\x*2,\y*2) -- (\x*2,\yy*2)         node{};  
    \fi

    \pgfmathparse{\x!= \maxwidth? int(1):int(0)}
    \ifnum\pgfmathresult>0
    \draw[blankedge] (\x*2,\y*2) to (\xx*2,\y*2)         node{};   
    \fi
  }
}
\path (0,0) -- node[fillednode] {\Large\blank} (2,0);
\path (2,0) -- node[fillednode] {\Large\blank} (4,0);
\path (4,0) -- node[filledlabel] {\Large$t$} (4,2);
\path (4,2) -- node[filledlabel] {\Large$h$} (4,4);
\path (4,4) -- node[fillednode] {\Large\blank} (6,4);
\path (6,4) -- node[fillednode] {\Large\blank} (8,4);
\path (8,4) -- node[filledlabel] {\Large$e$} (8,6);
\path (8,6) -- node[fillednode] {\Large\blank} (10,6);
\path (10,6) -- node[fillednode] {\Large\blank} (12,6);
\path (12,6) -- node[filledlabel] {\Large$s$} (12,8);
\path (12,8) -- node[fillednode] {\Large\blank} (14,8);
\path (14,8) -- node[filledlabel] {\Large$i$} (14,10);
\path (14,10) -- node[fillednode] {\Large\blank} (16,10);
\path (16,10) -- node[filledlabel] {\Large$s$} (16,12);
\path (16,12) -- node[fillednode] {\Large\blank} (18,12);
\path (18,12) -- node[fillednode] {\Large\blank} (20,12);

  \begin{pgfonlayer}{background}
    \draw[ultra thick,red, opacity=1] (0,0) -- (2,0) -- (4,0) -- (4,2) -- (4,4) -- (6,4) -- (8, 4) -- (8, 6) -- (10, 6) -- (12, 6) -- (12, 8) -- (14, 8) -- (14, 10) -- (16, 10) -- (16, 12) -- (18, 12) -- (20, 12);
  \end{pgfonlayer}

\node[finalnode] at (\maxwidth*2,\maxheight*2) {};

\draw[latex-latex, ultra thick] (\maxwidth*2 + 0.5,-0.5) node[right]{\Huge$\encoderlength$} -- (-0.5,-0.5) -- (-0.5,\maxheight*2 + 0.5) node[above]{\Huge$\tokenlength$};
\end{tikzpicture}
}
\caption[Label topology]{Unrolled label topology with allowed vertical transitions,
with a highlighted path for the sequence
$\mathcal{A}(\text{``\blank\blank{}th\blank\blank e\blank\blank s\blank i\blank s\blank \blank''}) = \text{``thesis''}$.
The terminal node is marked in the top-right corner.}
\label{fig:lattice_rnnt}
\end{figure}

Our \emph{decoder} model defines the probability distribution
over labels $\alpha_u$ as
\begin{align*}
p_u(\alpha_u \mid \smalldots) & :=
\begin{cases}
p_u(\Delta t_u {=} 1 \mid \smalldots), & \alpha_u = \blank, \\
p_u(\Delta t_u {=} 0 \mid \smalldots) \cdot q_u(\alpha_u \mid \smalldots) , & \alpha_u \in \Sigma
\end{cases}
\\
p_u( \Delta t_u \mid \smalldots) & :=
\begin{cases}
\sigma(-\operatorname{FF}_{\operatorname{emit}}( z_u^{\operatorname{fast}} )), & \Delta t_u = 1 \\
\sigma(\phantom{-}\operatorname{FF}_{\operatorname{emit}}(z_u^{\operatorname{fast}})), &
\Delta t_u = 0 
\end{cases}
\\
q_u(\alpha_u \mid \smalldots) &:=
\operatorname{softmax}_{\Sigma}
( \operatorname{FF}_{\Sigma}( z^{\text{fast}}_{u} )), \quad \alpha_u \in \Sigma
\\
z^{\text{fast}}_{u} & :=
\operatorname{Readout}(h_{t_u}, z^{\text{slow}}_{s_{u}}) \\
z^{\text{slow}}_{s_{u}} & := \operatorname{SlowRNN} (y_1^{s_{u} - 1})
\end{align*}
where $\sigma$ is the sigmoid function
and $\Delta t \in \{0,1\}$,
where $\Delta t_u = 0$ means that we emit a new non-blank label ($\Delta s_u = 1$)
and $\Delta t_u = 1$ means that we proceed forward in the time dimension
without emitting a non-blank label ($\Delta s_u = 0$).
Thus $\Delta t_u = 1$ can be understand as a reinterpretation of the blank label $\blank$.
Then we have a separate probability distribution $q$
over the labels $\Sigma$ (excluding $\blank$).
$\operatorname{FF}_{\operatorname{emit}}$, $\operatorname{FF}_{\Sigma}$
are linear transformations,
$\operatorname{Readout}$ is a linear transformation with maxout activation,
and $\operatorname{SlowRNN}$ is an LSTM.

\section{Training}

The loss is defined as
\[ L := -\log p(y_1^S \mid x_1^{T'}) = -\log \sum_{\alpha_1^U : y_1^S} p(\alpha_1^U \mid x_1^T) . \]
As we use the simplified transducer model
where $\operatorname{Readout}$ does not depend on $\alpha_{u-1}$,
we can efficiently calculate the exact full sum over all alignments $\alpha_1^U : y_1^S$
and do not need the maximum approximation \cite{zeyer2020:transducer}.

We use zoneout \cite{krueger2017zoneout} for the $\operatorname{SlowRNN}$
and optionally recurrent weight dropout \cite{wan13dropconnect} for the encoder BLSTMs.

We use the Adam optimizer \cite{kingma2015adam}
with learning rate scheduling based on cross validation scores.
Additionally, we reset the learning rate back to the initial value after a larger number of epochs,
after the model already converged,
and start over with the learning rate scheduling.
We train for 128 epochs. 
The long amount of training had a huge effect on the overall performance.

\subsection{Pretraining}

We use a pretraining scheme
where we schedule multiple aspects of the training:
\begin{itemize}
\item We grow the encoder from 3 layers with 500 dim.~%
up to 6 layers with 1000 dimensions \cite{zeyer2018:attanalysis}.
\item We increase the dropout rates.
\item
We use curriculum learning and start with shorter sequences initially.
\item We use linear learning rate warmup from $0.0001$ to $0.001$.
\item We use a higher initial time reduction factor 20 in the encoder
and reduce it to the final factor 6.
\end{itemize}

\subsection{Distributed Multi-GPU Training}

Our distributed training implementation
uses independent trainer (worker) instances per GPU.
Each worker independently loads the dataset.
To make sure that every worker uses a different part of the dataset,
it is common to use striding.
Striding has the disadvantage that it is very IO intensive
in this setting where every worker loads the dataset independently
and often becomes the bottleneck.
So we came up with the idea to use a different random seed
for the shuffling of the dataset for every worker,
to replace the striding.
This greatly improved the IO in our case
and made the training much faster.

Additionally, every worker independently trains an own copy of the model
for multiple update steps, until the models get synchronized
by averaging the parameters over all workers.
We made the further improvement that we do not synchronize
after a fixed number of steps,
but instead after a fixed time interval.
A fixed number of steps implies that the training is always as slow
as the slowest worker,
and variations in the runtime often lead to some workers
being slower than others even on same hardware.
Synchronizing after a fixed time interval does not have this problem,
while being more stochastic.

We synchronize only after 100 seconds
to reduce the communication between workers.
The workers can potentially be on different computing nodes
and might need to communicate over network,
which can result in 1-2 seconds for the synchronization.
We train on either 8 or 16 GPUs.

\section{Decoding \& Language Model Combination}

Our beam search decoding tries to find the sequence $\hat{y}_1^{\hat{S}}$
given $x_1^{T'}$ which maximizes the probability,
i.e.~specifically
\begin{align*}
x_1^{T'} \mapsto \hat{S}, \hat{y}_1^{\hat{S}} & := \argmax_{S, y_1^S} \log p(y_1^S \mid x_1^{T'}) \\
& \approx \mathcal{A} \circ \argmax_{U, \alpha_1^U} \log p(\alpha_1^U \mid x_1^{T'})
\end{align*}
We perform alignment-synchronous decoding,
i.e.~all hypotheses are in the same alignment step $u$
when being pruned \cite{zeyer2020:transducer,saon2020rnnt}.
We merge hypotheses by summing their scores
when they correspond to the same word sequence after BPE-merging.

The training recipe for our BPE-10K LSTM LM \cite{irie2020:phd}
has been adapted for the new BPE-1k label set
but otherwise no changes have been made.
%
%
\emph{Shallow fusion} (SF) \cite{gulcehre2016monolingual} is a log-linear combination
of the $\log$-scores of external LM and ASR model scores during the recognition process,
with scale $\lmscale$ for the LM
and scale $\labelscale$ for the acoustic (non-blank) label probability $q$,
while we do not add an own scale for $p(\Delta t)$.
Specifically, we use the score
\begin{align*}
\log p^{\text{SF}}_u(\alpha_u \mid \smalldots) & :=
\begin{cases}
\log p_u(\Delta t_u {=} 1 \mid \smalldots), & \alpha_u = \blank, \\
\log p_u(\Delta t_u {=} 0 \mid \smalldots) \\
		\quad \phantom{x} + \labelscale \cdot \log q_u (\alpha_u \mid \smalldots) \\
		\quad \phantom{x} + \lmscale \cdot \log p_{\operatorname{\small LM}}( \alpha_u \mid \smalldots )
		, & \alpha_u \in \Sigma
\end{cases} .
\end{align*}
We experimented with fixing the label scale at $\labelscale=1$ or $\labelscale=1 - \lmscale$.


Inspired by \cite{mcdermott2019density,variani2020hat,meng2021ilm}
we also tried to \emph{subtract the internal LM} log score.
It assumes that we can factorize our model into a language and acoustic model.
Although our model is not directly formulated as such,
we can approximate the internal language model.
For that we used the estimated score $\log p_{\operatorname{\small ILM}}$
as shown in \cref{sec:internal_lm_estimation},
where we use the average of encoder features in the time dimension.
%
%
%
\begin{align*}
\log p^{\text{SF-ILM}}_u(\alpha_u \mid \smalldots) & :=
\begin{cases}
\log p_u(\Delta t_u {=} 1 \mid \smalldots), & \alpha_u = \blank, \\
\log p_u(\Delta t_u {=} 0 \mid \smalldots) \\
		\phantom{.} + \labelscale \cdot \log q_u (\alpha_u \mid \smalldots) \\
		\phantom{.} + \lmscale \cdot \log p_{\operatorname{\small LM}}( \alpha_u \mid \smalldots ) \\
		\phantom{.} - \ilmscale \cdot \log p_{\operatorname{\small ILM}}( \alpha_u \mid \smalldots )
		, & \alpha_u \in \Sigma
\end{cases}
\end{align*}

%


\subsection{Internal LM Estimation}
\label{sec:internal_lm_estimation}

The transducer is trained on audio-text pairs
but learns an implicit prior model on the text.
This is explicitly given by the context dependency on previous labels.
In this transducer case, the \decoder{} is also explicitly modeled
such that it models the most important part of this prior
as it operates only on the text-only part
and runs label-synchronous.
This prior is an implicit internal LM in our acoustic model
\[ p_{\mathrm{prior}}( y ) = \sum_{x} p_{\mathrm{AM}}(y \mid x) \cdot p(x) \]
which can not be calculated efficiently in general.
To approximate the internal LM,
we replace the encoder input
to the rest of the model ($\operatorname{Readout}$).
%
%
%
We either use a $0$ vector or the encoder mean (\ilmAvgName).
The mean is computed over the time dimension for each sequence separately.%
\footnote{We also tested several other variants but got mixed inconclusive results.
In another work \cite{zeineldeen2021ilm}, we investigate variants on the ILM estimation in more detail
for attention-based encoder-decoder models.}

We evaluate the estimated internal LM on
text-only data.
%
In \cref{tab:internal_lm_estimation}, the BPE-level perplexities (PPL)
are shown and compared against the LSTM LM which was
trained only on text data,
but without any overlap to the audio transcriptions \cite{panayotov2015librispeech}.

\begin{table}[t]
\centering
\caption[Internal LM estimation on Librispeech]{Perplexity and WER measurements on Librispeech dev-other of a transducer model.
Note that the BPE-level (1k units) perplexity is evaluated without the EOS token, since the transducer has no explicit end-of-sequence symbol.
Compared are both setting the encoder ($h$) to $0$ and to the mean over the time-dimension (\ilmAvgName).
The LSTM and Trafo-LM are trained on text-only data without overlap to
the audio transcriptions \cite{panayotov2015librispeech}.
}
\label{tab:internal_lm_estimation}
\begin{tabular}{|c|c|c|c|c|}
\hline
\multirow{2}{*}{Model} & \multirow{2}{*}{Epochs} & \multicolumn{2}{c|}{Perplexity} & WER\\
 &  & $h=0$ & \ilmAvgName & [\%]\\
\hline\hline
\multirow{5}{*}{Transducer BPE-1K} & 8 & 82.76 & 67.47 & 36.41\\
\cline{2-5}
 & 16 & 49.32 & 38.89 & 17.16\\
\cline{2-5}
 & 32 & 45.13 & 32.86 & 11.85\\
\cline{2-5}
 & 64 & 46.53 & 31.94 & \phantom{0}9.69\\
\cline{2-5}
 & 133 & 47.05 & 31.37 & \phantom{0}8.92\\
\hline
LSTM LM & 20 & \multicolumn{2}{c|}{15.40} & $-$\\
\hline
Trafo LM & 39 & \multicolumn{2}{c|}{14.44} & $-$\\
\hline
\end{tabular}
\end{table}

\subsection{EOS Modelling}
\label{sec:lm_eos_modelling}

In contrast to language models or attention models,
transducers and models with explicit time modeling
do not have to model the end-of-sentence/sequence explicitly
with an additional token (denoted as \eos).
Instead the search ends when all input frames have been consumed.
However, for LM integration, when only considering actual output symbols,
the information about when the sequence should end is not considered.
This additional information is usually ignored in the literature,
however it provides valuable information to the search process.





Our approach is to combine the LM EOS~probability
with $p_u(\Delta t {=} 1)$ ($\blank$) in the last time frame ($t_u = T$)
because that determines the EOS in the transducer.
\begin{align*}
&\log p^{\text{SF-ILM+EOS}}_u(\alpha_u \mid \smalldots) \\
& \phantom{x} :=
\begin{cases}
\log \blankproblibrilhs{},
& \aligncur = \blank, t_{\alignpos-1} < T \\
\emitscaleeos \log \blankproblibrilhs{} \\
\phantom{x} + \lmscaleeos \log p_{\operatorname{LM}}(\eos \mid \smalldots),
& \aligncur = \blank, t_{\alignpos-1} = T\\
\log \emitproblibrilhs{} \\
\phantom{x} + \labelscale \cdot \log q_u (\alpha_u \mid \smalldots) \\
\phantom{x} + \lmscale \cdot \log p_{\operatorname{\small LM}}( \alpha_u \mid \smalldots ) \\
\phantom{x} - \ilmscale \cdot \log p_{\operatorname{\small ILM}}( \alpha_u \mid \smalldots ) ,
& \aligncur \in \Sigma
\end{cases}
\end{align*}
%
%
Usually 
 $\emitscaleeos = \lmscaleeos = 0.5$ yielded good performance,
although it was not tuned properly.

\section{Experiments}

We perform experiments on LibriSpeech \cite{panayotov2015librispeech}.
Our model training and decoding is implemented in RETURNN \cite{zeyer2018:returnn},
based on TensorFlow \cite{tensorflow2015}.
The distributed multi-GPU training is implemented with Horovod \cite{sergeev2018horovod}.
We make use of Mingkun Huang's warp-transducer loss implementation%
\footnote{\scriptsize\url{https://github.com/HawkAaron/warp-transducer}}.
Our decoder uses the builtin RETURNN features for stochastic variables
and searches over $\alpha_u$.
This uses GPU-based batched one-pass decoding with the external LM and internal LM subtraction.
We publish all the configuration files needed to reproduce the experiments%
\footnote{\scriptsize\url{https://github.com/rwth-i6/returnn-experiments/tree/master/2021-transducer}}.

\begin{figure}[t]
\begin{center}
\includegraphics[width=\columnwidth]{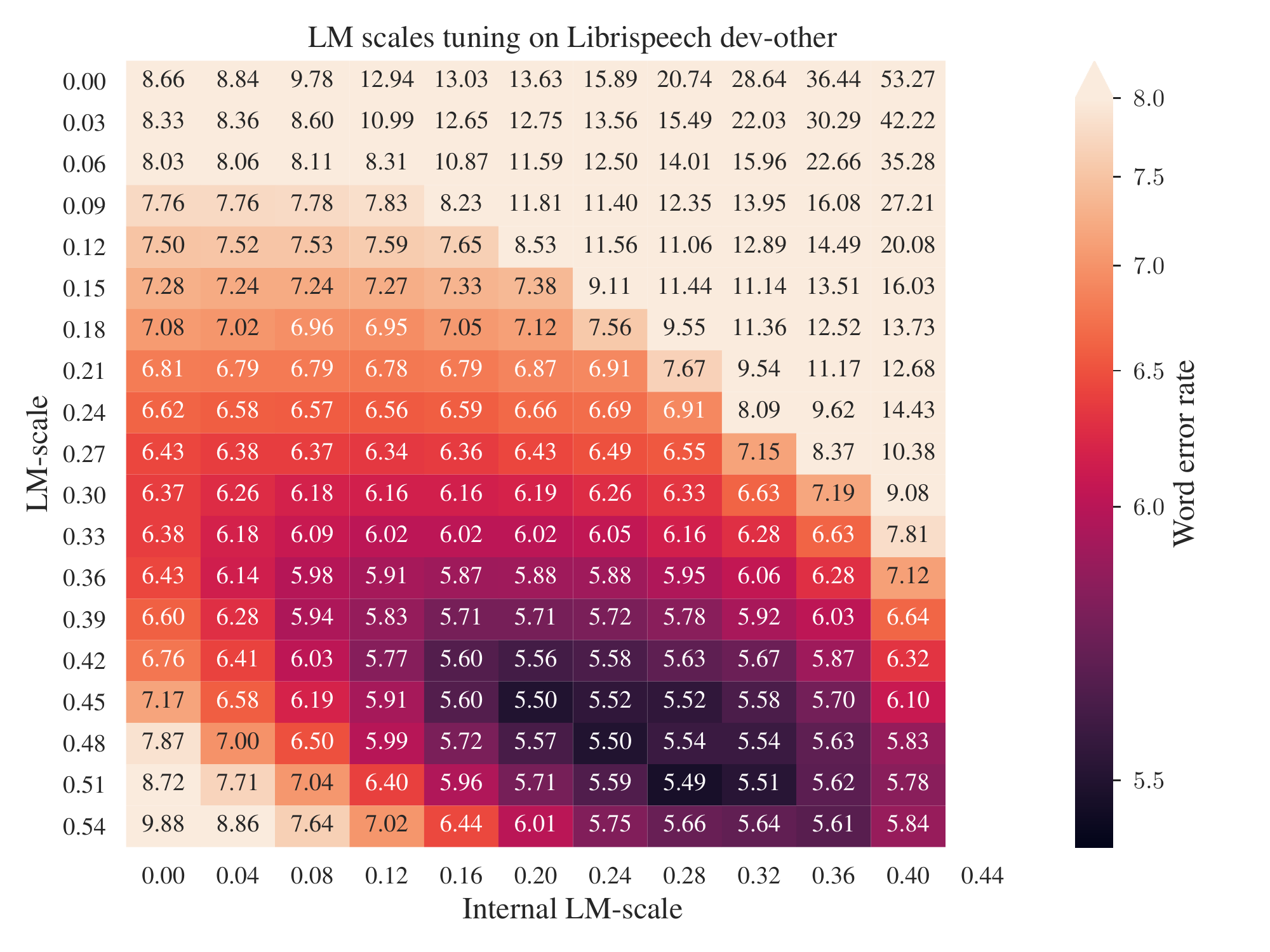}
\vspace{-8mm}
\end{center}
\caption[Tuning of LM scales on Librispeech dev-other]{Tuning of LM scales of a transducer  
with EOS-modeling.
The baseline without an external LM has $8.66$\% WER on dev-other with a beam size of 24.
$\labelscale=1-\lmscale$. 
}
\label{fig:librispeech_lm_scales}
\end{figure}

We have a variety of different exponent scales for our
log-linear modeling, as well as additional parameters for EOS-modeling.
Label scale $\labelscale$ which is set to either $\labelscale=1$ or $\labelscale=1-\lmscale$,
the emission model scale $\emitscale$, and the scales for external and internal LM \lmscale, \ilmscale,
respectively.
Additionally for EOS-modeling $\emitscaleeos$ and $\lmscaleeos$ is used, although it was fixed to $\emitscaleeos=\lmscaleeos=0.5$.
The scaling factors $\lmscale$ and $\ilmscale$ have to be tuned jointly
on a held-out dataset, as can be seen in \cref{fig:librispeech_lm_scales},
with $\labelscale=1-\lmscale$. 
They were tuned separately for each subset dev-clean and dev-other.
Results for LM integration are presented in \cref{tab:libri_lm_integration}
and in \cref{tab:libri_lm_integration_eos} with additional EOS-modeling.
With shallow fusion of just the LM we already see a significant WER improvement
by over 22\% relative. 
When additionally subtracting the internal LM,
a further significant improvement is observed
by over 14\% relative over the shallow fusion. 
The \ilmAvgName ILM estimation seems to be better than \ilmZeroName
except on test-other.
The effect of EOS gives us 7\% relative improvement. 
We also test a stronger Transformer LM in \cref{tab:libri_lm_integration_eos}
(perplexities in \cref{tab:internal_lm_estimation})
and see further improvement.

\begin{table}[t]
\centering
\caption[External LM integration results using different methods]{
We investigate the effect of LM integration for the model
with either shallow fusion (SF) or additional internal LM (ILM) subtraction.
All experiments were conducted with a fixed \textbf{beam-size 24} and
\textbf{without EOS-modeling}, the LSTM-LM has a BPE-1K level perplexity of $15.4$ on dev-other.}
\setlength{\tabcolsep}{0.25em}
\begin{tabular}{|c|c|c|c|c|c|c|}
\hline
\multirow{3}{*}{LM} & \multirow{3}{*}{\shortstack{LM\\Integration\\Method}} &  \multirow{3}{*}{\shortstack{Label scale\\\labelscale}}& \multicolumn{4}{c|}{WER [\%]}\\
&  &  & \multicolumn{2}{c|}{{dev}} & \multicolumn{2}{c|}{{test}}\\
& &  & {clean} & {other} & {clean} & {other}\\
\hline\hline
\textemdash & \textemdash & \multirow{2}{*}{$\labelscale=1\phantom{{} - \lmscale}$} & 3.22 & 8.76 & 3.30 & 8.70 \\

\cline{1-2}\cline{4-7}
\multirow{3}{*}{LSTM} & \multirow{2}{*}{SF} & &
2.53 & 6.79 & 2.66 & 6.99 \\


\cline{3-7}
&  & \multirow{2}{*}{$\labelscale=1-\lmscale$} &
2.47 & 6.50 & 2.57 & 6.70 \\

\cline{2-2}\cline{4-7}
& SF-ILM(\ilmAvgName) &  &
\textbf{2.29} & \textbf{5.63} & \textbf{2.36} & \textbf{6.39}\\
\hline
\end{tabular}
\label{tab:libri_lm_integration}
\end{table}

\begin{table}
\centering
\caption[External LM integration results with EOS modeling]{
We investigate the effect of LM integration for the model 
with either shallow fusion (SF) or additional internal LM (ILM) subtraction.
All experiments were conducted with a fixed \textbf{beam-size 24} and \textbf{EOS-modeling} (last blank frame), the LSTM-LM has a BPE-1K level perplexity of $15.4$ on dev-other.
In \cref{fig:librispeech_lm_scales} the heat map for the joint tuning over
$\lmscaleother$ and $\ilmscaleother$.
$\emitscaleeos = \lmscaleeos = 0.5$.
}
\setlength{\tabcolsep}{0.25em}
\begin{tabular}{|c|c|c|c|c|c|c|}
\hline
\multirow{3}{*}{LM} & \multirow{3}{*}{\shortstack{LM\\Integration\\Method}} &
\multirow{3}{*}{\shortstack{Label scale\\\labelscale}} & \multicolumn{4}{c|}{WER [\%]}\\
&  &  & \multicolumn{2}{c|}{{dev}} & \multicolumn{2}{c|}{{test}}\\
& & & {clean} & {other} & {clean} & {other}\\
\hline\hline
\textemdash & \textemdash & $\labelscale=1\phantom{{}-\lmscale}$ & 3.20 & 8.66 & 3.28 & 8.60\\
\hline\hline
\multirow{3}{*}{LSTM} & \multirow{2}{*}{SF} & $\labelscale=1\phantom{{}-\lmscale}$&
2.52 & 6.69 & 2.65 & 6.85\\

\cline{3-7}
& & \multirow{2}{*}{$\labelscale=1-\lmscale$} &
2.45 & 6.35 & 2.55 & 6.68\\

\cline{2-2}\cline{4-7}
& SF-ILM(\ilmAvgName) &  &
\textbf{2.26} & \textbf{5.49} & \textbf{2.42} & \textbf{5.91}\\
\hline\hline

\multirow{3}{*}{Trafo} & SF & \multirow{3}{*}{$\labelscale=1-\lmscale$} &
2.41 & 6.29 & 2.52 & 6.56\\

\cline{2-2}\cline{4-7}

& SF-ILM(\ilmAvgName) & &
\textbf{2.17} & \textbf{5.28} & \textbf{2.23} & {5.74} \\
\cline{2-2}\cline{4-7}

& SF-ILM(\ilmZeroName) & &
{2.22} & {5.32} & {2.25} & \textbf{5.60} \\

\hline
\end{tabular}
\label{tab:libri_lm_integration_eos}
\end{table}

\subsection{Error Analysis}

One of the sources of errors when looking at an entire system
are errors the model made when its prediction was wrong.
We look at the percentages of substitution, deletion, and insertion errors of the word error rate (WER).
Especially interesting is the comparison between different models and their respective LM integration.
Also of interest are how long the hypothesized sentences are, relative to the reference transcription.
The transducer seems to model the hypothesis length better than hybrid (without rescoring) and attention-based models, although adding an external LM seems to help the attention model.
Overall we can see that introducing the external LM helps with substitutions and insertion errors,
while the deletions actually increase.
In comparison to the attention-based model,
the transducer model has significantly less insertion errors,
but more deletion errors, relative to the overall WER.


\begin{table}[t]
\caption{We investigate the different type of word errors on various models.
Hybrid \cite{luescher2019:librispeech}, Attention \cite{zeyer2019:trafo-vs-lstm-asr},
and Transducer (ours).
With either shallow fusion (SF) or additional internal LM (ILM) subtraction.
For log-linear combination in the transducer case, $\labelscale=1-\lmscale$.}
\centering
\setlength{\tabcolsep}{0.2em}
\begin{tabular}{|c|c|c|S[table-format=2.2]|S[table-format=2.2]|S[table-format=2.2]|c|}
\hline
\multirow{2}{*}{Model} & \multirow{2}{*}{LM} & \multirow{2}{*}{\shortstack{LM\\integration}} & \multicolumn{3}{c|}{Edit operations [\%]} & WER\\
     & & &   {Sub.} &   {Del.} &  {Ins.} &   [\%] \\
\hline\hline
   Attention & None  & \textemdash  &        80.40 &        6.96 &       12.65 &     9.93 \\
\cline{2-7}

             & LSTM & SF &            79.42 &         5.68 &        14.90 &      7.50 \\
\hline\hline
Hybrid  & 4-gr. &  SF &  75.75 &       12.98 &       11.27 &     9.37 \\
\hline\hline

\multirow{5}{*}{Transd.} & None & \textemdash &              82.52 &         7.55 &         9.93 &      8.76 \\
\cline{2-7}

& \multirow{3}{*}{LSTM} & SF &           79.10 &        10.93 &         9.97 &      6.50 \\
\cline{3-7}

&   & SF-ILM(\ilmAvgName)  &         80.90 &         9.06 &        10.04 &      5.63 \\
\cline{3-7}

 &  & \multirow{2}{*}{\shortstack{SF-ILM(\ilmAvgName)\\+EOS}}  &         79.81 &        10.15 &        10.04 &      5.49 \\
\cline{2-2}\cline{4-7}

     & Trafo&   &         80.45 &        9.55 &        10.00 &      5.28 \\

\hline
\end{tabular}
\end{table}

%
%

\section{Conclusions \& Future Work}

The subtraction of the ILM helped to improve the model by a lot (over 14\% relative)
over the already strong shallow fusion.
The EOS modeling also helped (7\% relative).
We noticed that all recognition experiments are very sensitive to the LM/ILM scales.
The long training time also had a huge effect on the final performance.

As future work,
we plan to study the effect of the label unit and to test simple characters and other subword variations,
similar to \cite{zeineldeen20:phon-att}.
The encoder model might get improvements by more recent advancements \cite{gulati2020conformer}.
The decoder can be extended as well \cite{zeyer2020:transducer}.
We also can potentially improve the ILM estimation.
Finally, we expect to get improvements by min.~WER training.

\section{Acknowledgements}


We thank Yingbo Gao for providing us with the Transformer LM
on our BPE-1k label set.
This project has received funding from the European Research Council (ERC)
under the European Union’s Horizon 2020 research and innovation programme
(grant agreement n\textsuperscript{o}~694537, project "SEQCLAS"). The work
reflects only the authors' views and the European Research Council
Executive Agency (ERCEA) is not responsible for any use that may be made
of the information it contains.
This work was partly funded by the Google Focused Award "Pushing the
Frontiers of ASR: Training Criteria and Semi-Supervised Learning".
%

\bibliographystyle{IEEEtran}
\bibliography{transducer}

\end{document}